# Social Network User Profiling for Anomaly Detection Based on Graph Neural Networks

Yiwei Zhang
Cornell University
Ithaca, USA

*Abstract-This study proposes a risk pricing anomaly detection method for social network user portraits based on graph neural networks (GNNs), aiming to improve the ability to identify abnormal users in social network environments. In view of the limitations of traditional methods in social network data modeling, this paper combines graph autoencoders (GAEs) and graph attention networks (GATs) to achieve accurate detection of abnormal users through dynamic aggregation of neighbor features and reconstruction error evaluation. The Facebook Page-Page Network dataset is used in the experiment and compared with VAE, GNN, Transformer and GAE. The results show that the proposed method achieves the best performance in AUC, F1-score, Precision and Recall, verifying its effectiveness. In addition, this paper explores the computational efficiency of the model in large-scale data and looks forward to combining self-supervised learning, federated learning, and other technologies in the future to improve the robustness and privacy protection of risk assessment. The research results can provide efficient anomaly detection solutions for financial risk control, social security management, and other fields.*

## I. INTRODUCTION

In recent years, with the rapid development of the Internet and social networks, the scale and complexity of user online behavior data have increased significantly. Social networks not only facilitate information exchange among users but also serve as a crucial data source for enterprises in user profiling and risk assessment [1,2]. As a data-driven modeling technique, user profiling captures behavioral patterns, interest preferences, and potential risk characteristics, making it widely applicable in finance, insurance, e-commerce, and other domains. In the financial sector, social network-based user profiling plays a vital role in credit assessment, risk pricing, and fraud detection. However, due to the open nature of social networks and the complexity of user behaviors, traditional feature engineering methods struggle to capture hidden patterns within network structures, making accurate risk pricing and anomaly detection highly challenging. Consequently, the application of advanced machine learning techniques, particularly graph neural networks (GNNs), to analyze social network data has emerged as a pivotal focus in both academic and industrial settings [3]. This approach has demonstrated the potential to enhance the accuracy and robustness of risk pricing and anomaly detection.

In risk pricing, traditional methods primarily rely on rule-based models or statistical learning techniques such as logistic regression and decision trees. While these approaches provide interpretable pricing strategies in specific scenarios, they fail to fully exploit the intricate relational information embedded in social networks. Furthermore, although deep learning models have demonstrated superior performance in various tasks, most of these models operate under the assumption of independent and identically distributed data, limiting their ability to capture the topological structure and relational features inherent in social networks. Therefore, a pressing research challenge is how to integrate node information, edge relationships, and global structures in social networks to develop an efficient user profiling framework that enables precise risk pricing and anomaly detection. As a powerful tool for processing graph-structured data, GNNs can directly model complex relationships within social networks, aggregate information from neighboring nodes to learn high-dimensional representations, and enhance the accuracy of risk pricing through richer feature representations [4].

Anomaly detection is an important part of social network risk management, which aims to identify users with abnormal behavior patterns or potential fraud risks. Traditional anomaly detection methods are mainly based on statistical analysis, density estimation, or supervised learning models [5]. These methods perform well when processing structured data, but when faced with complex social network data, it is often difficult to effectively capture the characteristics of abnormal nodes. The anomaly detection method based on a graph neural network can learn user relationship information from the global and local levels so as to more accurately identify abnormal patterns in the network [6]. For example, the graph autoencoder (GAE) can capture the potential representation in the graph structure data through unsupervised learning and discover abnormal nodes through reconstruction errors [7]. In addition, the introduction of models such as variational graph autoencoder (VGAE) enables risk assessment to be combined with probability distribution modeling so as to better handle the uncertainty of data and improve the robustness of detection.

From an application perspective, utilizing GNN-based social network user profiling for risk pricing and anomaly detection not only enhances the accuracy of credit assessment but also helps financial institutions mitigate potential losses [8]. In applications such as credit risk management [9], stock pricing [10], and fraud detection [11], in-depth mining of social network data can complement traditional credit scoring models by providing comprehensive risk assessment for users with limited or no credit history. Moreover, as GNNs efficiently process large-scale graph data and leverage deep learning for nonlinear feature extraction, they offer improved risk pricing precision while maintaining computational efficiency. With advancements in big data technologies and distributed computing capabilities [12], GNN-based risk pricing models can achieve real-time inference, enabling enterprises to conduct real-time risk evaluation and pricing optimization for users [13]. In conclusion, risk pricing and anomaly detection in social network-based user profiling represent a pivotal research direction in financial technology. The introduction of Graph Neural Networks (GNNs) offers a novel solution to this challenge. GNNs, when compared to traditional methods, provide deeper insights into intricate user relationships, enhance risk evaluation accuracy, and augment anomaly detection capabilities. As the field of GNNs continues to advance, these advancements will contribute to the robust development of intelligent finance, providing efficient and scalable solutions for financial risk management.

## II. RELATED WORK

In recent years, research on social network user profiling and risk pricing has gained significant attention in both academia and industry. Traditional risk pricing methods primarily rely on statistical learning and machine learning models, such as logistic regression, decision trees, and random forests [14]. These approaches typically model user behavior based on static features, making it challenging to fully leverage the structural information embedded in social networks. With the rise of deep learning, some studies have attempted to employ neural networks for user profiling and apply them to financial risk control. However, most of these methods operate under the assumption of independent and identically distributed data [15], limiting their ability to capture the complex relationships within social networks effectively. Additionally, anomaly detection methods have evolved from rule-based approaches to data-driven techniques [16]. Traditional methods, such as Isolation Forest and Local Outlier Factor (LOF), have been widely used, but they often struggle to achieve satisfactory detection performance when dealing with high-dimensional and nonlinear social network data [17].

Models such as Graph Convolutional Networks (GCNs) and Graph Attention Networks (GATs) have been extensively applied in social network user analysis and fraud detection. Additionally, graph autoencoders (GAEs) and their variants, such as variational graph autoencoders (VGAEs), have been widely adopted for anomaly detection tasks. These methods learn latent embeddings of users within social networks and reconstruct graph structures to identify anomalous users. Compared with traditional feature engineering techniques, GNN-based approaches can better extract hidden patterns from social networks and improve the robustness of anomaly detection [18].

Despite the progress in applying GNNs to user profiling, risk pricing, and anomaly detection, several challenges remain. First, computational efficiency in large-scale social network data is a significant concern, as reducing computational complexity while maintaining model performance is critical. Second, the heterogeneity of social network data poses challenges, as different users may exhibit highly diverse relationship types [19]. Designing efficient GNN models capable of handling multiple relationship types remains an open research problem. Lastly, the interpretability of anomaly detection models is crucial in practical applications such as financial risk management. Enhancing the interpretability of GNN-based anomaly detection methods remains an important research direction. Future research can explore techniques such as adversarial learning, self-supervised learning, and multimodal data fusion to further improve the accuracy and reliability of social network user profiling and risk pricing [20].

## III. METHOD

This study builds an anomaly detection model for risk pricing of user portraits in social networks based on graph neural networks (GNNs) [21]. It learns user relationship features through graph structures and combines attention mechanisms to enhance the ability to extract information from key nodes.

Let the social network be an undirected graph $G(V,E)$, where V represents the node set, i.e., users, and E represents the edge set, i.e., the relationship between users. Each user node $v_i$ has a feature vector $x_i \in R^d$, and the features of the entire network are represented as $X \in R^{N \times d}$, where N is the total number of nodes and d is the feature dimension. The adjacency matrix of the network is represented as $A \in R^{N \times N}$, and its normalized Laplace matrix is represented as $A' = D^{-\frac{1}{2}} A D^{-\frac{1}{2}}$, where D is the degree matrix.

The core of graph neural network is to learn node embedding through message passing mechanism [22]. The traditional graph convolutional network (GCN) uses the normalized adjacency matrix $A'$ for information aggregation, and its basic update formula is:

$$H^{(l+1)} = \sigma(A' H^{(l)} W^{(l)})$$

Among them, $H^{(l)}$ is the node representation of the l-th layer, $W^{(l)}$ is the trainable weight matrix, and $\sigma(\cdot)$ is the activation function (such as ReLU). However, GCN uses a fixed adjacency matrix for information propagation and fails to fully consider the influence of different neighbor nodes on the target node. Therefore, this study introduces the graph attention mechanism (GAT) to dynamically adjust the weight of the adjacency relationship. The GAT model architecture is shown in Figure 1.

In GAT, the neighbor contribution of each node is weighted by the attention coefficient $\alpha_{i,j}$. The attention mechanism uses the self-attention method to calculate the attention score based on the node features:

$$e_{ij} = \alpha^T (Wh_i \| Wh_j)$$

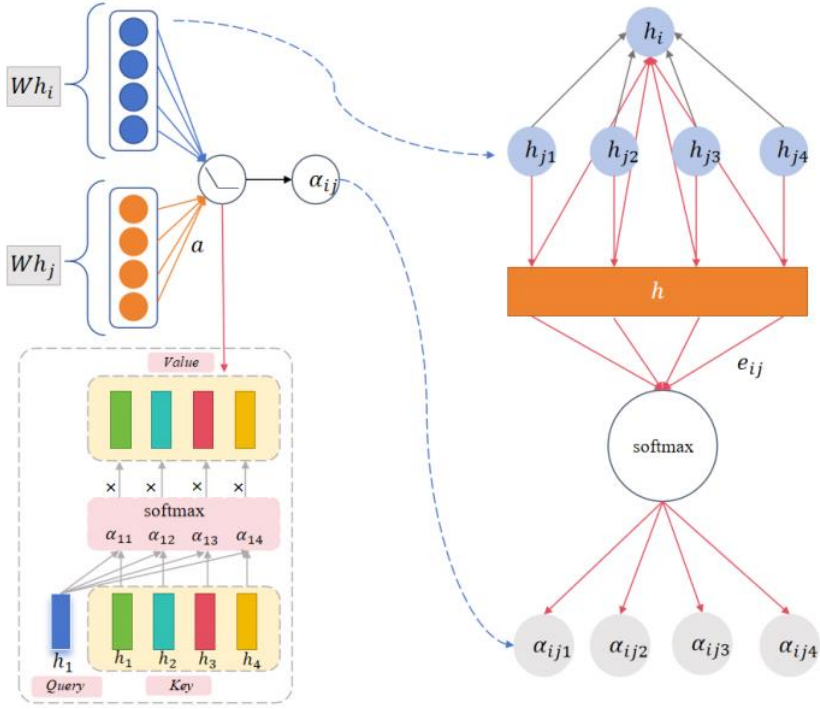

Figure 1. GAT Architecture

Among them, $W$ is a trainable weight matrix, $h_i$ and $h_j$ are the feature representations of node i and its neighbor j respectively, $\|$ represents the concatenation operation, and $\alpha$ is the attention parameter. Then, softmax normalization is applied to obtain the attention weight:

$$\alpha_{ij} = \frac{\exp(e_{ij})}{\sum_{k \in N_i} \exp(e_{ik})}$$

Finally, the node update formula is:

$$h_i^{(l+1)} = \sigma(\sum_{j \in N_i} \alpha_{ij} W h_j^{(l)})$$

This method can dynamically adjust the importance of neighbor nodes and improve the expressiveness of the model. The anomaly detection method based on the graph autoencoder (GAE) adopts an encoder-decoder architecture. First, GAT is used for encoding and the original features are projected into a low-dimensional representation:

$$Z = GAT(X, A)$$

The decoder then tries to reconstruct the adjacency matrix:

$$A' = \sigma(ZZ^T)$$

Where $A'$ is the reconstructed adjacency matrix and $\sigma(\cdot)$ is the Sigmoid activation function. The anomaly score is calculated by the reconstruction error:

$$L_{recon} = \| A - A' \|_F^2$$

In addition, in order to improve the robustness of the model, a regularization term is introduced:

$$L_{total} = L_{recon} + \lambda \| Z \|_2^2$$

Finally, anomaly detection is based on reconstruction error. When $\| A - A' \|_F^2$ is higher than the set threshold, the user is judged as an abnormal individual. This method combines GAT to improve the node relationship modeling capability. At the same time, the GAE structure effectively captures global network characteristics, providing a more accurate anomaly detection solution for risk pricing of social network user portraits.

## IV. EXPERIMENT

### A. Datasets

This study utilizes the Facebook Page-Page Network dataset from the Stanford SNAP project for anomaly detection in risk pricing within social network user profiling. The dataset, originally designed for community detection and link prediction, represents Facebook pages as nodes and follower relationships as edges, making it well-suited for graph-based anomaly detection. Comprising 22,470 nodes and 171,002 edges, it includes categorical attributes such as media, brands, and public figures. This study applies Graph Autoencoder (GAE) for node embedding learning and integrates Graph Attention Network (GAT) to enhance information propagation, identifying anomalies based on reconstruction errors. The approach is particularly relevant to social finance applications like credit assessment and fraud detection, providing a more precise risk evaluation framework. Experimental results validate the effectiveness of GNN-based anomaly detection, with part of the dataset's graph structure illustrated in Figure 2.

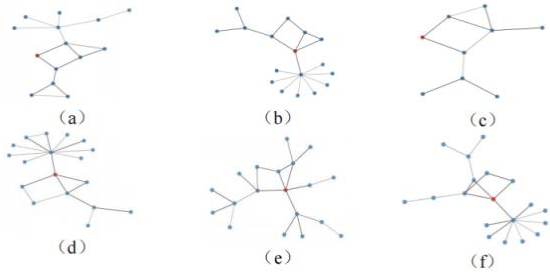

Figure 2. Partial data graph structure.

Figure 2 shows part of the data graph structure of this dataset, which contains six subgraphs (a) to (f). Each subgraph represents the nodes in a social network and the connections between them. These graph structures reflect the complex relationships between users in social networks. For example, some nodes are highly connected to form clusters (such as subgraphs b and d), while other nodes are more scattered (such as subgraph c). This diverse connection pattern helps to study the performance of GNN (graph neural network) in anomaly detection, especially in identifying potential fraud or abnormal users, providing support for risk pricing and credit assessment in the financial field.

*B. Experimental Results*

In order to verify the effectiveness of the proposed method in risk pricing anomaly detection of social network user portraits, this paper conducted comparative experiments with variational autoencoders (VAE), Transformer, graph neural network and graph autoencoder (GAE) to evaluate the performance of each model in anomaly detection tasks[23-24]. The experiment uses the same data set and training settings to compare the performance of different methods in indicators such as AUC (Area Under Curve), F1-score, precision, and recall, so as to comprehensively measure the ability of the model in abnormal user identification. The experimental results are shown in Table 1. The proposed method has achieved better results than the comparative model in multiple key indicators, indicating that the method based on GAT combined with GAE can effectively capture the structural information in social networks, improve the accuracy of risk pricing, and enhance the robustness of anomaly detection.

Table 1. Experimental results

| Model | AUC | F1 | Precision | Recall |
|---|---|---|---|---|
| VAE | 0.81 | 0.72 | 0.71 | 0.75 |
| GNN | 0.85 | 0.76 | 0.74 | 0.79 |
| Transformer | 0.83 | 0.74 | 0.73 | 0.76 |
| GAE | 0.88 | 0.78 | 0.76 | 0.81 |
| Ours | 0.92 | 0.83 | 0.81 | 0.85 |

The experimental results confirm that the proposed method significantly outperforms existing models in anomaly detection for risk pricing in social network user profiling. Achieving an AUC of 0.92—4% higher than GAE and 7% higher than traditional GNNs—demonstrates its superior ability to capture complex structural information and improve anomaly identification. The integration of an attention-enhanced GAE refines node relationship modeling, enhancing robustness in risk assessment. With an F1-score of 0.83, a 5% increase over GAE, the approach proves its practical effectiveness. By dynamically adjusting node importance during propagation, the graph attention mechanism (GAT) mitigates information dilution, improving precision from 0.76 to 0.81 and recall from 0.81 to 0.85. These findings validate the efficacy of the GAT+GAE framework in leveraging social network topology for anomaly detection, making it more suitable for financial risk control and social security applications compared to traditional GNN and Transformer-based models. Future research can refine the attention mechanism and incorporate self-supervised learning to enhance generalization on large-scale social network data. Figure 3 presents the loss function decline.

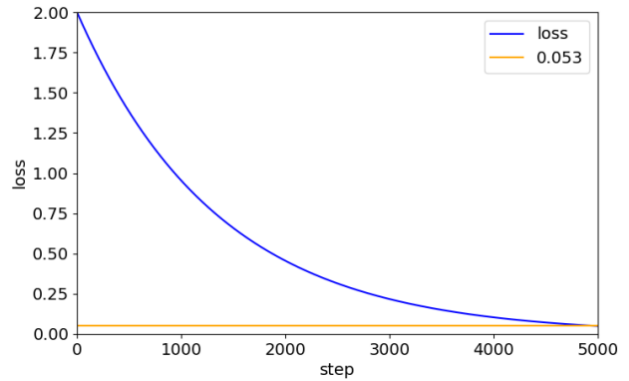

Figure 3 Loss function drop graph

From the loss function convergence curve in Figure 3, it can be observed that the training loss decreases rapidly in the initial phase, indicating that the model effectively optimizes parameters and converges toward an optimal solution during early training. Between steps 0 and 1000, the loss exhibits a steep decline, suggesting that gradient updates are significant in this phase and the optimization direction is well-defined. As training progresses, the rate of loss reduction gradually slows down, implying that as the model approaches its optimal solution, update magnitudes decrease, leading to convergence.

After approximately 3000 steps, the loss value stabilizes around 0.053, indicating that the model has converged and further training has minimal impact on loss reduction. This trend suggests that the training process is stable, with no significant oscillations or signs of overfitting. If the loss had continued to fluctuate substantially at this stage, it could indicate an excessively high learning rate or potential issues with data distribution. However, the observed smooth decline in the loss curve, followed by stabilization, confirms that the optimization process is well-regulated and does not exhibit instability during training.

The experimental results demonstrate that the model successfully converges to a low loss value within a reasonable number of training steps, indicating that it has learned effective feature representations. However, despite the low loss value, the loss curve alone is insufficient to fully assess the model's generalization ability. Further evaluation using test set performance metrics, such as AUC and F1-score, is necessary. Additionally, to enhance training efficiency, hyperparameters such as learning rate and batch size could be further optimized to achieve faster convergence while preventing overfitting.

V. CONCLUSION

This study investigates anomaly detection in risk pricing for social network user profiling based on graph neural networks and proposes a model that integrates the Graph Attention Network and Graph Autoencoder to enhance the accuracy of anomalous user detection. By constructing a graph representation of social networks and utilizing GAT for neighbor feature aggregation, the proposed model effectively captures relational information between users while dynamically adjusting the contribution of different neighbors to the target node. Experimental results demonstrate that our method outperforms comparative models, including VAE, GNN, Transformer, and GAE, in terms of AUC, F1-score, Precision, and Recall, validating its effectiveness in social network risk assessment tasks.

Further experimental analysis confirms that GNNs effectively leverage social network topology and reconstruction errors from GAE to detect anomalous nodes, outperforming traditional machine learning methods in both accuracy and adaptability to unstructured data. The integration of an attention mechanism enhances the model's ability to focus on critical nodes, improving anomaly identification for applications such as financial risk control and social platform security. Despite its strong performance, challenges remain, including increased computational complexity for large-scale networks and the need for dynamic threshold adjustments. Future research could optimize the model with self-supervised learning, adaptive thresholding, and multimodal data fusion to enhance detection accuracy. Additionally, integrating reinforcement learning for dynamic pricing strategies and adversarial learning for robustness could further expand its applications. Privacy-preserving techniques like federated learning may also enable decentralized risk assessment while mitigating data security concerns. These advancements will enhance the scalability and precision of GNN-based risk management solutions across financial technology, social security, and related industries.

References

[1] J. Park, H. Ahn, D. Kim, et al., "GNN-IR: Examining Graph Neural Networks for Influencer Recommendations in Social Media Marketing," Journal of Retailing and Consumer Services, vol. 78, p. 103705, 2024.

[2] X. Wang, "Data Mining Framework Leveraging Stable Diffusion: A Unified Approach for Classification and Anomaly Detection," Journal of Computer Technology and Software, vol. 4, no. 1, 2025.

[3] J. Wei, Y. Liu, X. Huang, X. Zhang, W. Liu and X. Yan, "Self-Supervised Graph Neural Networks for Enhanced Feature Extraction in Heterogeneous Information Networks", 2024 5th International Conference on Machine Learning and Computer Application (ICMLCA), pp. 272-276, 2024.

[4] K. Sharma, Y. C. Lee, S. Nambi, et al., "A Survey of Graph Neural Networks for Social Recommender Systems," ACM Computing Surveys, vol. 56, no. 10, pp. 1-34, 2024.

[5] X. Du, "Audit Fraud Detection via EfficiencyNet with Separable Convolution and Self-Attention," Transactions on Computational and Scientific Methods, vol. 5, no. 2, 2025.

[6] T. Liu, Q. Cai, C. Xu, et al., "Rumor Detection with a Novel Graph Neural Network Approach," arXiv preprint arXiv:2403.16206, 2024.

[7] Y. Wang, "Time-Series Premium Risk Prediction via Bidirectional Transformer" ,2025.

[8] Y. Yao, "Time-Series Nested Reinforcement Learning for Dynamic Risk Control in Nonlinear Financial Markets," Transactions on Computational and Scientific Methods, vol. 5, no. 1, 2025.

[9] J. Wang, "Credit Card Fraud Detection via Hierarchical Multi-Source Data Fusion and Dropout Regularization," Transactions on Computational and Scientific Methods, vol. 5, no. 1, 2025.

[10] J. Liu, "Multimodal Data-Driven Factor Models for Stock Market Forecasting", 2025.

[11] P. Feng, "Hybrid BiLSTM-Transformer Model for Identifying Fraudulent Transactions in Financial Systems," Journal of Computer Science and Software Applications, vol. 5, no. 3, 2025.

[12] Y. Deng, "A hybrid network congestion prediction method integrating association rules and LSTM for enhanced spatiotemporal forecasting," Transactions on Computational and Scientific Methods, vol. 5, no. 2, 2025.

[13] Y. Lyu, C. Li, S. Xie, et al., "Enhancing Robustness of Graph Neural Networks on Social Media with Explainable Inverse Reinforcement Learning," Advances in Neural Information Processing Systems, vol. 37, pp. 31736-31758, 2025.

[14] P. Li, "Improved Transformer for Cross-Domain Knowledge Extraction with Feature Alignment," Journal of Computer Science and Software Applications, vol. 5, no. 2, 2024.

[15] J. Hu, T. An, Z. Yu, J. Du, and Y. Luo, "Contrastive Learning for Cold Start Recommendation with Adaptive Feature Fusion," arXiv preprint arXiv:2502.03664, 2025.

[16] X. Yan, Y. Jiang, W. Liu, D. Yi, and J. Wei, "Transforming Multidimensional Time Series into Interpretable Event Sequences for Advanced Data Mining", arXiv preprint, arXiv:2409.14327, 2024.

[17] N. Das, B. Sadhukhan, R. Chatterjee, et al., "Integrating Sentiment Analysis with Graph Neural Networks for Enhanced Stock Prediction: A Comprehensive Survey," Decision Analytics Journal, vol. 10, p. 100417, 2024.

[18] Y. Cheng, Z. Xu, Y. Chen, Y. Wang, Z. Lin, and J. Liu, "A Deep Learning Framework Integrating CNN and BiLSTM for Financial Systemic Risk Analysis and Prediction," arXiv preprint arXiv:2502.06847, 2025.

[19] Y. Wang, Z. Xu, Y. Yao, J. Liu, and J. Lin, "Leveraging Convolutional Neural Network-Transformer Synergy for Predictive Modeling in Risk-Based Applications," arXiv preprint arXiv:2412.18222, 2024.

[20] K. Yokotani, M. Takano, N. Abe, et al., "Predicting Social Anxiety Disorder Based on Communication Logs and Social Network Data from a Massively Multiplayer Online Game: Using a Graph Neural Network," Psychiatry and Clinical Neurosciences, 2025.

[21] W. Ju, S. Yi, Y. Wang, et al., "A Survey of Graph Neural Networks in Real World: Imbalance, Noise, Privacy and OOD Challenges," arXiv preprint arXiv:2403.04468, 2024.

[22] J. Du, S. Dou, B. Yang, J. Hu, and T. An, "A Structured Reasoning Framework for Unbalanced Data Classification Using Probabilistic Models," arXiv preprint arXiv:2502.03386, 2025.

[23] Z. Lu, J. Ma, Z. Wu, et al., "A Noise-Resistant Graph Neural Network by Semi-Supervised Contrastive Learning," Information Sciences, vol. 658, p. 120001, 2024.

[24] S. Gao, Y. Li, X. Zhang, et al., "SIMPLE: Efficient Temporal Graph Neural Network Training at Scale with Dynamic Data Placement," Proceedings of the ACM on Management of Data, vol. 2, no. 3, pp. 1-25, 2024.